\documentclass{article}
\usepackage{booktabs}
\usepackage{adjustbox}
\usepackage{color}

\usepackage{CJKutf8}

\usepackage{tcolorbox}
\usepackage{tabularx} 
\definecolor{lightgray}{gray}{0.95}
\definecolor{purpleblue}{rgb}{0.5, 0.0, 0.5}

\usepackage[round]{natbib}
\usepackage[preprint]{neurips_2023}

\usepackage[utf8]{inputenc} 
\usepackage[T1]{fontenc}    
\usepackage{hyperref}       
\usepackage{url}            
\usepackage{booktabs}       
\usepackage{amsfonts}       
\usepackage{nicefrac}       
\usepackage{microtype}      
\usepackage{xcolor}         
\usepackage{authblk}
\title{E-EVAL: A Comprehensive Chinese K-12 Education Evaluation Benchmark for Large Language Models}

\author[1,2]{Jinchang Hou\thanks{Equal Contribution}}
\author[1,3]{Chang Ao\protect\footnotemark[1]}
\author[1,2]{Haihong Wu}
\author[1,2]{Xiangtao Kong}
\author[1,3]{\mbox{Zhigang Zheng}}
\author[4]{Daijia Tang}
\author[5]{Chengming Li}
\author[5]{Xiping Hu}
\author[6]{Ruifeng Xu} 
\author[1]{Shiwen Ni\thanks{Corresponding author}}
\author[1]{\mbox{Min Yang}\protect\footnotemark[2]}
\affil[1]{Shenzhen Institute of Advanced Technology, Chinese Academy of Sciences, Shenzhen, China}
\affil[2]{University of Science and Technology of China, Hefei, China}
\affil[3]{Southern University of Science and Technology, Shenzhen, China}
\affil[4]{UNION INFORMATION, Shenzhen, China}
\affil[5]{Shenzhen MSU-BIT University, Shenzhen, China}
\affil[6]{Harbin Institute of Technology (Shenzhen), Shenzhen, China}
\affil[ ]{\textit {\{sw.ni, min.yang, c.ao, zg.zheng\}@siat.ac.cn, \{jinchangh, taotaotao\}@ustc.edu, haihongw@mail.ustc.edu.cn, tangdj@szlhxx.com, licm@smbu.edu.cn, xuruifeng@hit.edu.cn}}
\begin{document}
\maketitle
\begin{CJK}{UTF8}{gbsn}
\begin{abstract}
With the accelerating development of Large Language Models (LLMs), many LLMs are beginning to be used in the Chinese K-12 education domain. The integration of LLMs and education is getting closer and closer, however, there is currently no benchmark for evaluating LLMs that focuses on the Chinese K-12 education domain. Therefore, there is an urgent need for a comprehensive natural language processing benchmark to accurately assess the capabilities of various LLMs in the Chinese K-12 education domain. To address this, we introduce the E-EVAL, the first comprehensive evaluation benchmark specifically designed for the Chinese K-12 education field. The E-EVAL consists of 4,351 multiple-choice questions at the primary, middle, and high school levels across a wide range of subjects, including Chinese, English, Politics, History, Ethics, Physics, Chemistry, Mathematics, and Geography. We conducted a comprehensive evaluation of E-EVAL on advanced LLMs, including both English-dominant and Chinese-dominant models. Findings show that Chinese-dominant models perform well compared to English-dominant models, with many scoring even above the GPT 4.0. However, almost all models perform poorly in complex subjects such as mathematics. We also found that most Chinese-dominant LLMs did not achieve higher scores at the primary school level compared to the middle school level. We observe that the mastery of higher-order knowledge by the model does not necessarily imply the mastery of lower-order knowledge as well. Additionally, the experimental results indicate that the Chain of Thought (CoT) technique is effective only for the challenging science subjects, while Few-shot prompting is more beneficial for liberal arts subjects. With E-EVAL, we aim to analyze the strengths and limitations of LLMs in educational applications, and to contribute to the progress and development of Chinese K-12 education and LLMs\footnote{\url{https://github.com/AI-EDU-LAB/E-EVAL}}\footnote{\url{https://huggingface.co/datasets/E-EVAL/E-EVAL}}\footnote{\url{https://eevalbenchmark.com}}.
\end{abstract}
\section{Introduction}
Large language models (LLMs) have made significant advancements in the field of natural language processing and artificial intelligence. The evaluation of the knowledge and reasoning capabilities embedded in these models has become progressively more difficult, leading to the development of multiple testing benchmarks. Novel benchmarks such as MMLU\citep{hendrycks2021measuring}, BIG-bench\citep{srivastava2022beyond}, and HELM\citep{liang2022holistic} span multiple domains and tasks, encompassing real-world examinations and textbook knowledge. These benchmarks evaluate not only language comprehension but also the models' ability in common sense reasoning, mathematical reasoning, and code generation. Concurrently, with the rapid development of Chinese LLMs, an increasing number of Chinese benchmarks have begun to surface. MMCU\citep{zeng2023measuring} focuses on professional domains, AGIEval\citep{zhong2023agieval} targets China's standardized tests, C-EVAL\citep{huang2023c} encompasses knowledge from middle school to professional fields, CMMLU\citep{li2023cmmlu} concentrates on Chinese culture and CMB\citep{wang2023cmb} focuses on the field of Chinese medicine. While these benchmarks primarily focus on the models' advanced ability, certain specific fields and topics crucial to the models might not receive adequate attention. Currently, there is no comprehensive assessment benchmark in the field of Chinese K-12 education that is important for assessing and analyzing the specifics of LLMs' learning of human knowledge at all stages.

\begin{figure*}[h]  
    \centering  
    \includegraphics[width=0.95\textwidth]{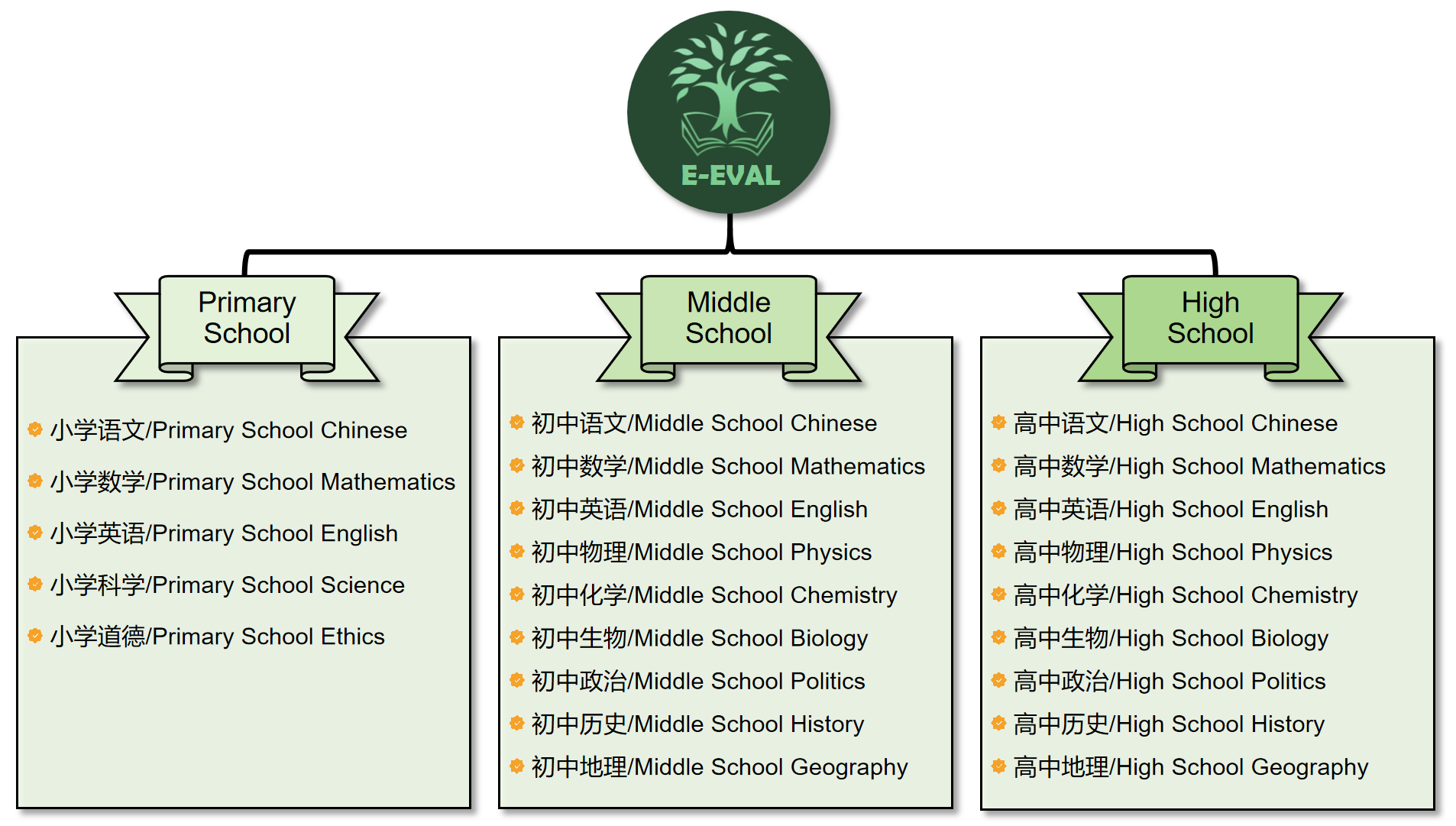}  
    \caption{Overview diagram of the E-EVAL benchmark.}  
    \label{f1}  
\end{figure*}

In this paper, we introduce E-EVAL, the first comprehensive evaluation suite focusing on Chinese K-12, aimed at evaluating basic models' knowledge and reasoning ability within the context of K-12. E-EVAL comprises 4,351 multiple-choice questions across primary, middle, and high school stages, as depicted in Figure \ref{f1} covering 23 subjects including \textit{Primary School Chinese, Primary School Mathematics, Primary School English, Primary School Science, Primary School Ethics, Middle School Chinese, Middle School Mathematics, Middle School English, Middle School Physics, Middle School Chemistry, Middle School Biology, Middle School Politics, Middle School History, Middle School Geography, High School Chinese, High School Mathematics, High School English, High School Physics, High School Chemistry, High School Biology, High School Politics, High School History, High School Geography}. We further categorize the questions into two types: arts and science, with science encompassing disciplines like Mathematics, Physics, Chemistry, and arts including Chinese, English, History, etc., noting that the difficulty of arts subjects is generally lower than that of science.

Our evaluation of multiple open-source and commercial models on E-EVAL reveals that Chinese-dominant models outperform English-dominant ones in Chinese K-12 performance, with several models even surpassing GPT 4.0. However, performance in complex subjects like Mathematics remains subpar. Additionally, we observed that advanced Chinese models struggle to achieve higher performance at lower educational levels compared to higher ones. This underscores the need for improvement in LLMs within the Chinese K-12 educational domain and highlights the potential value of E-EVAL as an important evaluation suite.

\begin{table}[ht]
  \begin{minipage}[t]{0.48\linewidth}
    \centering
    \begin{tabular}{lccc}
      \toprule
      \textbf{subject} & \textbf{\#Subject} & \textbf{\#Question} \\
      \midrule
      \multicolumn{3}{c}{\textbf{In terms of difficulty}} \\
      High School & 9 & 2225 \\
      Middle School & 9 & 1589 \\
      Primary School & 5 & 537 \\
      \midrule
      \multicolumn{3}{c}{\textbf{In terms of arts/science}} \\
      Arts & 13 & 2699 \\
      Science & 10 & 1652 \\
      \midrule
      \multicolumn{3}{c}{\textbf{In terms of split}} \\
      Dev & 23 & 115 \\
      Valid & 23 & 424 \\
      Test & 23 & 3812 \\
      \midrule
      Total & 23 & 4351 \\
      \bottomrule
    \end{tabular}
    \vspace{1em}
    \caption{Statistics of E-EVAL.}
    \label{tab:1statistics}
  \end{minipage}\hfill
  \begin{minipage}[t]{0.48\linewidth}
    \centering
    \begin{tabular}{cccc}
      \toprule
      \textbf{Option} & \textbf{E-EVAL} & \textbf{C-EVAL} & \textbf{MMLU}\\
      \midrule
      A & 24.3\% & 22.9\% & 23.1\% \\
      B & 26.1\% & 26.0\% & 24.7\% \\
      C & 25.8\% & 26.4\% & 25.5\% \\
      D & 23.8\% & 24.7\% & 26.7\% \\
      \bottomrule
    \end{tabular}
    \vspace{1em}
    \caption{Distribution of the correct answer.}
    \label{tab:distribution}
  \end{minipage}
\end{table}

\section{The E-EVAL Evaluation Benchmark}
\subsection{Design Principle}
E-EVAL is a meticulously crafted benchmark designed to evaluate the performance of LLMs within the diverse educational environments of Chinese K-12 education. It encompasses a comprehensive coverage of various subjects, categorizing them into arts and science to provide an all-encompassing evaluation across all educational stages. Overall statistics of E-EVAL are presented in Table \ref{tab:1statistics}, with detailed subject-wise data in Appendix \ref{app:statisticOfSbuject}.

The benchmark employs a multiple-choice question format,akin to \cite{hendrycks2021measuring},offering a clear and effective method for evaluating the precision and reasoning ability of LLMs. The questions, carefully selected and primarily sourced from homework and local small-scale exams, reflect the real educational setting while maintaining content originality and regional specificity. Special attention has been paid to the manual collection and fine processing of data, especially for subjects involving complex equations, to ensure high data integrity.

E-EVAL deliberately avoids using questions from national exams like the Gaokao to minimize the risk of data contamination, instead opting for mock tests and specific high school online exams. The choice to use PDF and Word documents as primary sources of information, rather than plain text or structured questions, further reduces the risk of data leakage. E-EVAL is extracted from thousands of test papers from different regions, schools, grades and subjects. This benchmark is particularly aimed at aiding developers in rapidly understanding and enhancing the capability of LLMs in processing subject-specific knowledge and content with cultural uniqueness in the context of Chinese education. Thus, E-EVAL is a comprehensive, effective, and culturally benchmark, intended to foster the advancement and optimization of LLMs in the educational technology field. 
\subsection{Data Collection}
\textbf{Subjects: }E-EVAL encompasses a range of subjects pivotal to Chinese K-12 education, covering key disciplines at primary, middle, and high school levels to cater to the educational needs of different age groups. This comprehensive coverage ensures E-EVAL's high applicability and representativeness across various educational stages. For better organization of these subjects, they have been divided into two main categories: arts and science, to more aptly reflect the nature and characteristics of each discipline. In the arts category, subjects that study unique aspects of human society, such as politics, economics, and culture, are included. These subjects not only focus on the transmission of knowledge but also emphasize the cultivation of thinking ability and humanistic literacy. This category encompasses subjects like Chinese, English, Politics, History, and Geography, covering the fields of social science and humanity. On the other hand, the science category includes natural science, applied science, and mathematical logic, emphasizing the cultivation of scientific methods and experimental skills, as well as understanding of the natural world and technological domains. This category covers subjects such as Mathematics, Physics, Chemistry, and Biology, encompassing the STEM (Science, Technology, Engineering, and Mathematics) fields. In total, E-EVAL covers 23 different subjects, including both arts and science, as illustrated in Figure \ref{f1}. 

\textbf{Sources: }Our data primarily come from free, regional homework, practice questions, and mock exams available on the internet\footnote{Our raw data comes from \url{https://www.zxxk.com/} and \url{https://zujuan.xkw.com/}.}. These are typically provided by schools, educational institutions, or teachers to assist students in consolidating knowledge. Unlike public exams like the Gaokao or Zhongkao, these local homework and practice questions tend not to circulate widely, thereby posing a lower risk of contamination. This choice of data source ensures better originality and quality of data, enhancing the credibility of the assessment. Moreover, these regional questions better reflect the real educational environment and academic requirements, as they are closer to what students encounter in their daily learning.

\begin{figure*}[h]  
\begin{tcolorbox}[colback=lightgray]
实验室需配制一种强酸溶液500mL，c($H^+$)=2mol/L，下列配制方法可行的是：\par
\textcolor{blue}{Laboratory needs to prepare a strong acid solution of 500mL, c($H^+$)=2mol/L, the feasible preparation method is:}\par
A. 取100mL5mol/L$H_2SO_4$，加入400mL水。\par
\textcolor{blue}{A. Take 100mL of 5mol/L $H_2SO_4$, add 400mL of water.}\par
B. 取100mL5mol/L$H_2SO_4$，加水稀释至500mL。\par
\textcolor{blue}{B. Take 100mL of 5mol/L $H_2SO_4$, dilute with water to 500mL.}\par
C. 取100mL5mol/LHCl，加水稀释至500mL。\par
\textcolor{blue}{C. Take 100mL of 5mol/L HCl, dilute with water to 500mL.}\par
D. 取100mL5mol/L$HNO_3$，加水稀释至500mL。\par
\textcolor{blue}{D. Take 100mL of 5mol/L $HNO_3$, dilute with water to 500mL.}\par
答案：B\par
\textcolor{blue}{Answer: B}\par
详解：A．100mL5mol/L $H_2SO_4$，加入400mL水溶液的体积要小于500mL，无法计算浓度，A不符合题意； B．c($H^+$)==2mol/L，B符合题意； C．c($H^+$)==1mol/L，C不符合题意 ； D．c($H^+$)==1mol/L，D不符合题意； 故选B。\par
\textcolor{blue}{Explanation: A. The volume of the solution of 100mL 5mol/L $H_2SO_4$ with 400mL of water is less than 500mL, and the concentration cannot be calculated, so A does not meet the requirements; B. c($H^+$) == 2mol/L, B meets the requirements; C. c($H^+$) == 1mol/L, C does not meet the requirements; D. c($H^+$) == 1mol/L, D does not meet the requirements; Therefore, choose B.}
\end{tcolorbox}
\caption{A development example with explanations from E-EVAL. English translations are provided beneath the relevant Chinese text.}
\label{fig:dev_example}
\end{figure*}
\textbf{Processing: }The collected data are in various formats, primarily PDFs and Microsoft Word documents. For arts subjects like Chinese and English, we employ scripts to automatically parse PDF and Word documents for structured data. However, for science subjects with complex formulas, manual parsing is necessary, converting formulas into standard LaTeX format. Most of the collected questions follow a format of one question with four options, and questions with fewer than four options are discarded. For questions with more than four options, one incorrect option is removed. After format conversion, we conduct three rounds of manual checks: the first ensures no data duplication, the second verifies the correctness and completeness of formulas, and the third confirms the accuracy of answers. After checks, the order of options is deliberately adjusted to achieve a more balanced distribution of correct answers among options A, B, C, and D, with the aim of minimizing the potential impact of option bias within the model. The distribution of correct answer is shown in Table \ref{tab:distribution}.
A total of 4351 questions were collected, categorized into development, validation, and test sets across the 23 subjects. Additionally, we selected five representative questions with explanations to support Few-shot assessments. A representative example with explanations is illustrated in Figure \ref{fig:dev_example}. The final development set contains 115 questions, the validation set contains 424 questions, and the test set contains 3812 questions,as shown in Table \ref{tab:1statistics}.
\subsection{E-EVAL Arts and Science}
We divided E-EVAL into two separate benchmarks by subject, E-EVAL Arts and E-EVAL Science. \textbf{E-EVAL Arts} includes 13 subjects: \textit{primary school Chinese, primary school English, primary school ethics, middle school Chinese, middle school English, middle school politics, middle school history, middle school geography, high school Chinese, high school English, high school politics, high school history and high school geography}. \textbf{E-EVAL Science} consists of 10 subjects: \textit{Primary school Mathematics, Primary school Science, Middle School Mathematics, Middle School Physics, Middle School Chemistry, Middle School Biology, High School Mathematics, High School Physics, High School Chemistry, and High School Biology.}

\subsection{Evaluation}
\begin{figure*}[h]  
\begin{tcolorbox}[colback=lightgray]
以下是中国关于高中生物的单项选择题，请选出其中的正确答案。\par
\textcolor{blue}{The following are multiple-choice questions about biology of high school in China. Please select the correct answer.}\par
人体内含有多种多样的蛋白质，每种蛋白质( )。\par
\textcolor{blue}{The human body contains various proteins, and each protein ( ).}\par
A. 都含有21种氨基酸。\par
\textcolor{blue}{A. Contains 21 types of amino acids.}\par
B. 都是在细胞内发挥作用。\par
\textcolor{blue}{B. Functions within cells.}\par
C. 都能调节生物体的生命活动。\par
\textcolor{blue}{C. Regulates the life activities of organisms.}\par
D. 都具有一定的空间结构。\par
\textcolor{blue}{D. Has a certain spatial structure.}\par
答案：D\par
\textcolor{blue}{Answer: D}\par
...[another 4 examples without explanation]...\par
下列关于植物激素作用的说法，错误的是( )。\par
\textcolor{blue}{Among the following statements about the effects of plant hormones, the incorrect one is ( )}\par
A. 种子在即将成熟时遇到高温天气出现“穗上发芽”的现象与脱落酸含量降低有关。\par
\textcolor{blue}{A. The phenomenon of "germination on the ear" when seeds are about to mature in high temperatures is related to the decrease in abscisic acid content.}\par
B. 摘除棉花的顶芽可以促进侧芽的萌发，主要是因为顶芽摘除后侧芽处生长素含量降低。\par
\textcolor{blue}{B. Removing the apical bud of cotton can promote the germination of lateral buds, mainly because the content of auxin at the lateral bud decreases after the apical bud is removed.}\par
C. 啤酒生产中用赤霉素处理大麦种子，可以使大麦在不发芽的状态下就能产生淀粉酶。\par
\textcolor{blue}{C. Treating barley seeds with gibberellin in beer production can allow barley to produce amylase without germination.}\par
D. 生长素浓度升高到一定水平时，会促进乙烯的合成，乙烯又会反过来促进生长素合成。\par
\textcolor{blue}{D. When the concentration of auxin rises to a certain level, it promotes the synthesis of ethylene, and ethylene, in turn, promotes the synthesis of auxin.}\par
答案：\textcolor{red}{D}\par
\textcolor{blue}{Answer: \textcolor{red}{D}}\par
\end{tcolorbox}
\caption{An example with five-shot-ao from E-EVAL. The red part is the response from model, English translations are provided beneath the relevant Chinese text.}
\label{fig:few-shot-ao}
\end{figure*}

\begin{figure*}[h]  
\begin{tcolorbox}[colback=lightgray]
以下是中国关于高中数学的单项选择题，请选出其中的正确答案。\par
\textcolor{blue}{The following are multiple-choice questions about high school mathematics in China. Please select the correct answers.}\par
已知某教学大楼共有四层，每层都有东、西两个楼梯，则从一层到四层不同的走法种数为( )。\par
\textcolor{blue}{Given that a teaching building has four floors, each with east and west stairs, the number of different ways to go from the first floor to the fourth floor is ( ).} \par
A. 7 \par
B. 8 \par
C. 9 \par
D. 10 \par
答案: 让我们一步一步思考,根据题意，教学大楼共有四层，每层都有东、西两个楼梯，则从一层到二层，有2种走法，同理从二层到三层、从三层到四层也各有2种走法，则从一层到四层共有\(2 \times 2 \times 2 = 8\)种走法。所以答案是B。\par
\textcolor{blue}{Answer: Let's think step by step. According to the problem, the teaching building has four floors, and each floor has east and west stairs. There are 2 ways to go from the first floor to the second floor, similarly from the second floor to the third floor, and from the third floor to the fourth floor, each having 2 ways. Therefore, the total number of ways from the first floor to the fourth floor is \(2 \times 2 \times 2 = 8\),so the answer is B.}\par
... [another 4 examples with explanation] ...\par
若集合$A=\{0,1,2\}$, 则A有( )个真子集。\par
\textcolor{blue}{If the set $A=\{0,1,2\}$, then $A$ has ( ) proper subsets.} \par
A. 7 \par
B. 8 \par
C. 9 \par
D. 6 \par
答案: \par
\textcolor{red}{让我们一步一步思考:集合 $A=\{0,1,2\}$ 的真子集是指除了集合本身和空集之外的所有可能子集。}\par \textcolor{red}{1、空集是 
$A$的真子集。}\par\textcolor{red}{2、包含一个元素的子集：$\{0\}$, $\{1\}$, $\{2\}$。}\par \textcolor{red}{3、包含两个元素的子集：$\{0,1\}$, $\{0,2\}$, $\{1,2\}$。}\par\textcolor{red}{4、包含所有元素的子集：$\{0,1,2\}$。}\par
\textcolor{red}{因此，共有  $2^3 - 1 = 7$个真子集，答案是A：7}\par

\textcolor{blue}{Answer:}

\textcolor{red}{Let's consider step by step: A set $A=\{0,1,2\}$ has subsets, excluding the set itself and the empty set.}

\textcolor{red}{1. The empty set is a proper subset of $A$.}

\textcolor{red}{2. Subsets containing one element: $\{0\}$, $\{1\}$, $\{2\}$.}

\textcolor{red}{3. Subsets containing two elements: $\{0,1\}$, $\{0,2\}$, $\{1,2\}$.}

\textcolor{red}{4. Subset containing all elements: $\{0,1,2\}$.}

\textcolor{red}{Therefore, there are $2^3 - 1 = 7$ proper subsets, the answer is A:7}

\end{tcolorbox}
\caption{An example with five-shot-cot from E-EVAL. The red part is the response from model, English translations are provided beneath the relevant Chinese text.}
\label{fig:few-shot-cot}
\end{figure*}
We use accuracy as the metric for evaluation. To ensure fairness, only the answers to the training and validation sets were disclosed, keeping the test set answers private. This approach prevents the incorporation of E-EVAL data in pre-training datasets. Users are invited to submit their predictions for the test set on our website\footnote{\url{https://eevalbenchmark.com}} to ascertain their accuracy. The site maintains a public leaderboard, where users have the discretion to publish their model's results.

\section{Experiment}
Here we detail our experiments. We tested 15 advanced models on this benchmark, encompassing various sizes, language orientations, and stages (pretrained or fine-tuned). This in-depth analysis of their performance offers a reliable reference point for future research in this field.
\subsection{Setup}
The experimental design of E-EVAL aims to evaluate the performance of LLMs on E-EVAL. Both open-source and proprietary advanced LLMs were tested. These models were prompted to select the correct choice from a set of questions with four options (ABCD), and regular expressions were used to extract the model's selected response. Three evaluation methods were developed: zero-shot\citep{kojima2022large}, few-shot-answer-only (few-shot-ao), and few-shot-chain-of-thought (few-shot-cot)\citep{wei2022chain}, to deeply analyze the models' knowledge and reasoning ability.
\subsection{Prompt}
We introduced the following phrase before each question: "以下是中国关于[subject]考试的单项选择题，请选出其中的正确答案（Here is a multiple-choice question from China's [subject] examination. Please select the correct answer）" In the zero-shot evaluation, the question and options were presented directly after the prompt, without any prior examples or additional information, requiring the model to rely solely on its existing knowledge and understanding to respond. 

For the few-shot-ao evaluation, we appended five related questions without explanation from the development set. This method helps the model better understand and adapt to the current question by leveraging these prior examples, enhancing the model's ability to adapt to new tasks using a minimal number of examples. In the few-shot-cot evaluation, we further included explanations and the prompt "让我们一步一步思考(Let's think step by step)" building on the few-shot-ao approach. This is designed to encourage the model to demonstrate its step-by-step reasoning process in solving the question, rather than just providing the answer. This evaluation mode emphasizes the model's reasoning ability, making its approach to problem-solving more akin to human thought processes. At the end of each question, we added "答案是：(Answer:)" to present the model's final conclusion. The examples of few-shot-ao and few-shot-cot are shown in Figure \ref{fig:few-shot-ao} and Figure \ref{fig:few-shot-cot}, respectively.

\subsection{Models}
We assessed 15 models from different countries, organizations, and sizes, as shown in Table \ref{tab:models_info}. For commercial models, we assessed ChatGPT\citep{openai_chatgpt_2022}, GPT 4.0\citep{openai_gpt4_2023}, ERNIE-Bot and ERNIE-Bot 4.0\citep{zhang2019ernie}. For open-source models, we tested Qwen-72B/7B\citep{bai2023qwen}, Yi-34B/6B-Chat\citep{01AI_Yi_2023}, ChatGLM3-6B\citep{du2022glm,zeng2022glm}, Baichuan2-13B/7B-Chat\citep{yang2023Baichuan}, Chinese-LLaMA-2-13B and Chinese-Alpaca-2-13B\citep{Chinese-LLaMA-Alpaca}, among other general models. Additionally, we evaluated the EduChat series\citep{dan2023educhat}, focusing on China's education field, including Educhat-sft-002-13B-Baichuan, Educhat-sft-002-7B. Please refer to Appendix \ref{app:models} for details on each model.
\begin{table}[ht]
  \centering
  \begin{tabularx}{\textwidth}{lXXc}
    \toprule
    \textbf{Model} & \textbf{Creator} & \textbf{\#Parameters} & \textbf{Access} \\
    \midrule
    Qwen-72B & Alibaba & 72B & Weights \\
    Ernie-Bot 4.0 & Baidu & undisclosed & API \\
    Yi-34B-Chat & 01.AI & 34B & Weights \\
    Ernie-Bot & Baidu & undisclosed & API \\
    GPT 4.0 & OpenAI & undisclosed & API \\
    Yi-6B-Chat & 01.AI & 6B & Weights \\
    ChatGLM3-6B & Tsinghua & 6B & Weights \\
    Qwen-7B & Alibaba & 7B & Weights \\
    Baichuan2-13B-Chat & Baichuan & 13B & Weights \\
    Baichuan2-7B-Chat & Baichuan & 7B & Weights \\
    ChatGPT & OpenAI & undisclosed & API \\
    Chinese-Alpaca-2-13B & HFL & 13B & Weights\\
    Educhat-sft-002-13B & ECNU & 13B & Weights \\
    Chinese-LLaMA-2-13B & HFL & 13B & Weights\\
    Educhat-sft-002-13B-Baichuan & ECNU & 13B & Weights \\
    \bottomrule
  \end{tabularx}
  \vspace{1em}
    \caption{Models evaluated in this paper.}
  \label{tab:models_info}
\end{table}
\subsection{Main Results}
The testing results of various models are presented in Table \ref{tab:model_performance}, where we report the average accuracy for three types of prompts across different categories. The accuracy for each of the three prompts is shown in Appendix \ref{app:threeprompts}. Detailed accuracy information for each subject is provided in Appendix \ref{app:resofsubjects}. Among the large-scale models, Alibaba's Qwen-72B-Chat model achieves the highest accuracy rate, averaging 88.8\%, attributed to its extensive parameter size and high-quality Chinese corpus. Baidu's ERNIE-Bot 4.0 follows closely in second place, trailing Qwen-72B-Chat by only 3.3 percentage points. Yi-34B-Chat and ERNIE-Bot demonstrate comparable overall performance, with a mere 1 percentage point difference in average accuracy. Notably, GPT 4.0 and ChatGPT exhibit poor performance, ranking 5th and 11th in accuracy, respectively. The underperformance of the GPTs may be attributed to the training corpus, where the Chinese corpus constitutes a low percentage. Among models with parameters less than 10B, Yi-6B-Chat performs the best, ranking 6th, approaching the accuracy of GPT 4.0 but still trailing behind Yi-34B-Chat. This suggests that models with larger parameter sizes demonstrate enhanced knowledge and inference, aligning with expectations. Qwen-7B-Chat, Baichuan2-13B-Chat,ChatGLM3-6B, and Baichaun2-7B-Chat closely follow, outperforming ChatGPT due to their rich and high-quality Chinese corpus, despite their smaller size. Chinese-LLaMA exhibits subpar performance, primarily attributed to the inadequacy of Chinese training data. Although EduChat is primarily trained on an educational corpus, it underperformed in the evaluation, potentially due to its focus on reinforcing psychological and pedagogical theories.

Overall, Chinese-dominant models that have been trained on a wide range of Chinese corpora show excellent performance in this evaluation. In the same series, the large model outperformed the small model. In addition, all models performed much better in arts subjects than in science subjects. This is in line with our expectations, as text comprehension is a strong point of LLMs, while logical reasoning has been a weak point of LLMs.
\begin{table}[ht]
  \centering
    \begin{tabularx}{\textwidth}{lXXXXXc}
    
      \toprule
      \textbf{Model} & \textbf{Arts} & \textbf{Science} & \textbf{Primary} & \textbf{Middle} & \textbf{High} & \textbf{Average} \\
      \midrule
      Random & 25.0 & 25.0 & 25.0 & 25.0 & 25.0 & 25.0 \\
      Qwen-72B-Chat & 92.5 & 84.0 & 89.3 & 93.1 & 85.6 & 88.9 \\
      Ernie-Bot 4.0 & 90.8 & 78.6 & 87.3 & 89.6 & 82.1 & 85.5 \\
      Yi-34B-Chat & 82.4 & 69.5 & 79.6 & 83.1 & 71.7 & 76.9 \\
      Ernie-Bot & 81.7 & 68.2 & 78.7 & 80.8 & 71.6 & 75.9 \\
      GPT 4.0 & 75.4 & 64.2 & 81.9 & 76.8 & 70.6 & 70.6 \\
      Yi-6B-Chat & 74.7 & 61.1 & 71.3 & 76.1 & 63.1 & 68.8 \\
      Qwen-7B-Chat & 67.3 & 50.1 & 69.7 & 65.9 & 53.3 & 59.9 \\
      Baichuan2-13B-Chat & 65.3 & 47.8 & 69.2 & 65.1 & 49.8 & 57.8 \\
      ChatGLM3-6B & 61.9 & 51.9 & 60.0 & 65.0 & 51.8 & 57.6 \\
      Baichuan2-7B-Chat & 62.2 & 45.0 & 61.2 & 61.3 & 48.6 & 54.8 \\
      ChatGPT & 60.5 & 46.8 & 68.3 & 58.2 & 48.8 & 54.6 \\
      Chinese-Alpaca-2-13B & 53.6 & 36.6 & 51.4 & 46.7 & 38.9 & 43.3 \\
      Educhat-sft-002-13B & 41.8 & 28.9 & 39.9 & 39.9 & 32.7 & 36.3 \\
      Chinese-LLaMA-2-13B & 44.2 & 31.9 & 39.2 & 38.5 & 33.2 & 35.9 \\
      Educhat-sft-002-13B-Baichuan & 40.2 & 29.3 & 40.8 & 38.4 & 32.2 & 35.5 \\
      \bottomrule
    \end{tabularx}
    \vspace{1em}
  \caption{Accuracy of multiple models in different categories.}
  \label{tab:model_performance}
\end{table}
\subsection{Insight and Analysis}

\textbf{Are large language models better at arts or sciences?}
Observations from Table \ref{tab:model_performance} regarding model performance across different subject categories reveal that all models perform better in arts subjects than in science subjects. Arts subjects emphasize memory, interpretation, and language understanding, aligning well with the basic construction of the models, which includes language processing and pattern recognition ability. Science subjects, on the other hand, involve logical reasoning, quantitative analysis, and solving complex problems, requiring strong logical reasoning and mathematical processing ability from the models. Therefore, it is logical that models exhibit better performance in arts subjects.

\begin{figure*}[h]  
    \centering  
    \includegraphics[width=1\textwidth]{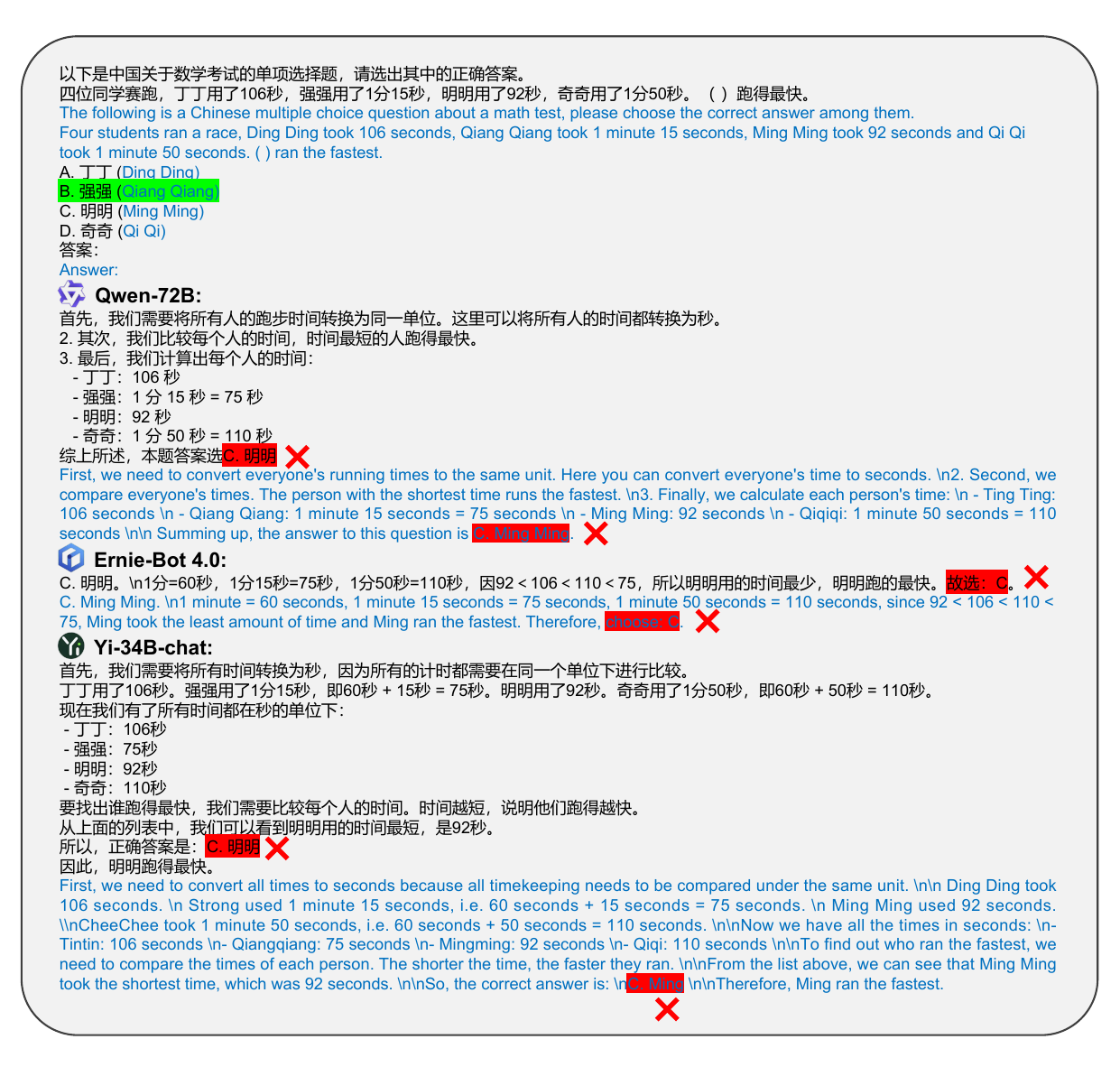}  
    \caption{A simple primary school math problem with the predictions of the top-3 models.}  
    \label{f-case1}  
\end{figure*}  
\begin{figure*}[h]  
    \centering  
    \includegraphics[width=1\textwidth]{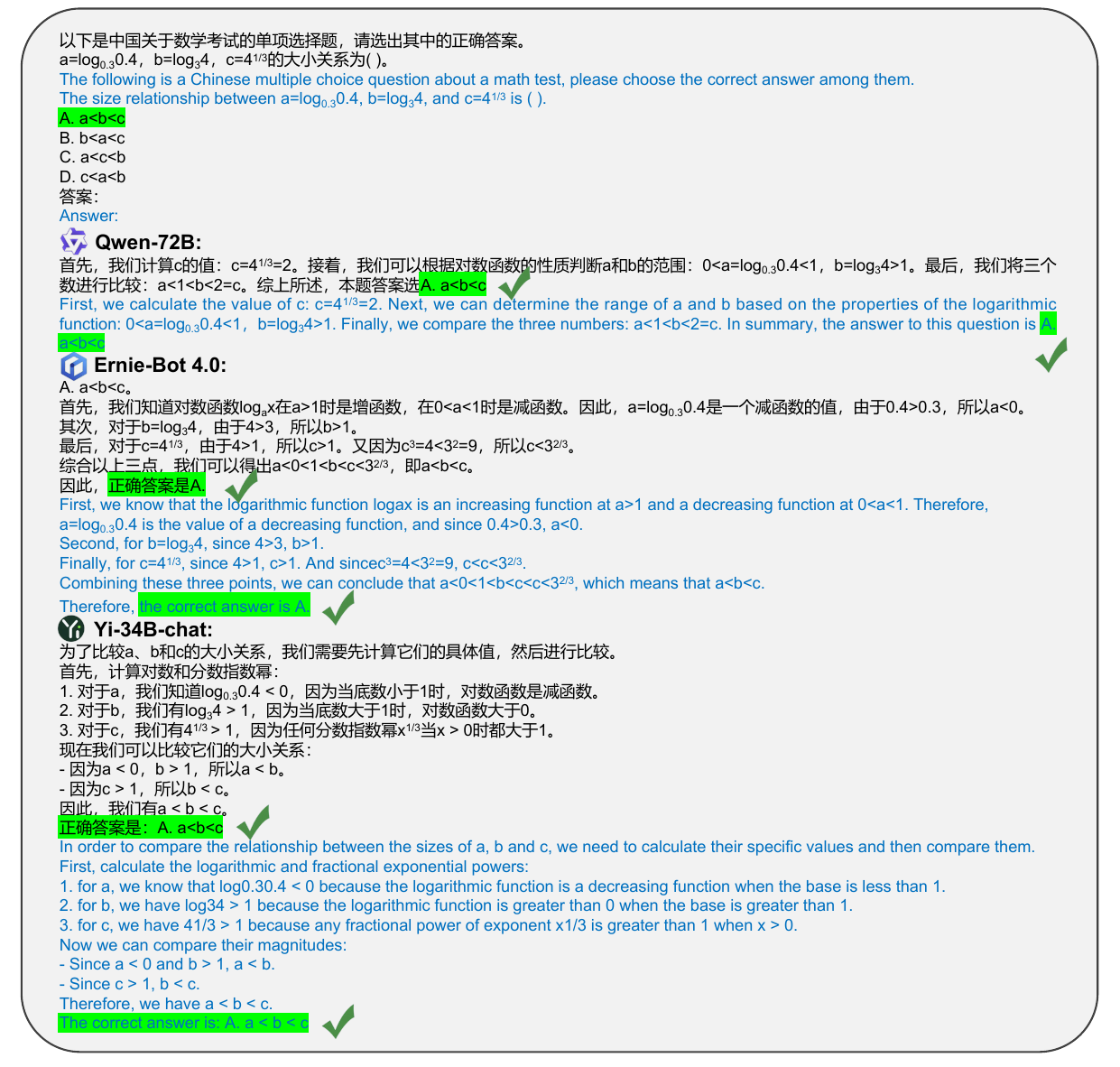}  
    \caption{A hard high school math problem with the predictions of the top-3 models.}  
    \label{f-case2}  
\end{figure*} 
\textbf{Does the simpler the question, the better the model performs?} 
Further observations of model performance at different educational levels in Table \ref{tab:model_performance} show that although the difficulty of the problems gradually increases from primary school to high school, the top-ranked models generally perform better at the middle school level than at the primary school level. This phenomenon is highly counter-intuitive because elementary school questions are far easier than middle school questions from a human cognitive perspective. As shown in Figure \ref{f-case1}, a very simple elementary school math question was answered incorrectly by the top three models on the E-EVAL leaderboard." \textit{Four students ran a race, Ding Ding took 106 seconds, Qiang Qiang took 1 minute 15 seconds, Ming Ming took 92 seconds and Qi Qi took 1 minute 50 seconds. ( ) ran the fastest.}". The correct answer is B: Qiang Qiang, but all three models predict C: Ming Ming. The Top-3 LLMs all thought that \textbf{92 seconds is faster than 75 seconds}. Ernie-Bot 4.0 model surprisingly generated such incredible results as \textbf{92 < 106 < 110 < 75}. This result seems to indicate that LLMs are not good at comparing sizes, yet we find that LLMs are capable of solving similarly complex high school math problems. As shown in the example in Figure 6, the LLMs can accurately determine the magnitude relationship between $\quad a=\log_{0.3}0.4,\quad b=\log_34,\quad c=4^{1/3}$ as $\log_{0.3}0.4<\log_34<4^{1/3}$.

We speculate that the pre-training data for these models is probable to use only middle and high school and college level knowledge and ignore the very simple elementary school level knowledge. It is possible that the developers believe that the primary school level is too simple, and that once more complex knowledge is mastered, the simpler knowledge will be automatically mastered. As a result, the model is trained with a bias toward solving higher stage topics and may perform poorly on simple knowledge that occurs less frequently in the training data.

\begin{figure*}[h]  
    \centering  
    \includegraphics[width=\textwidth]{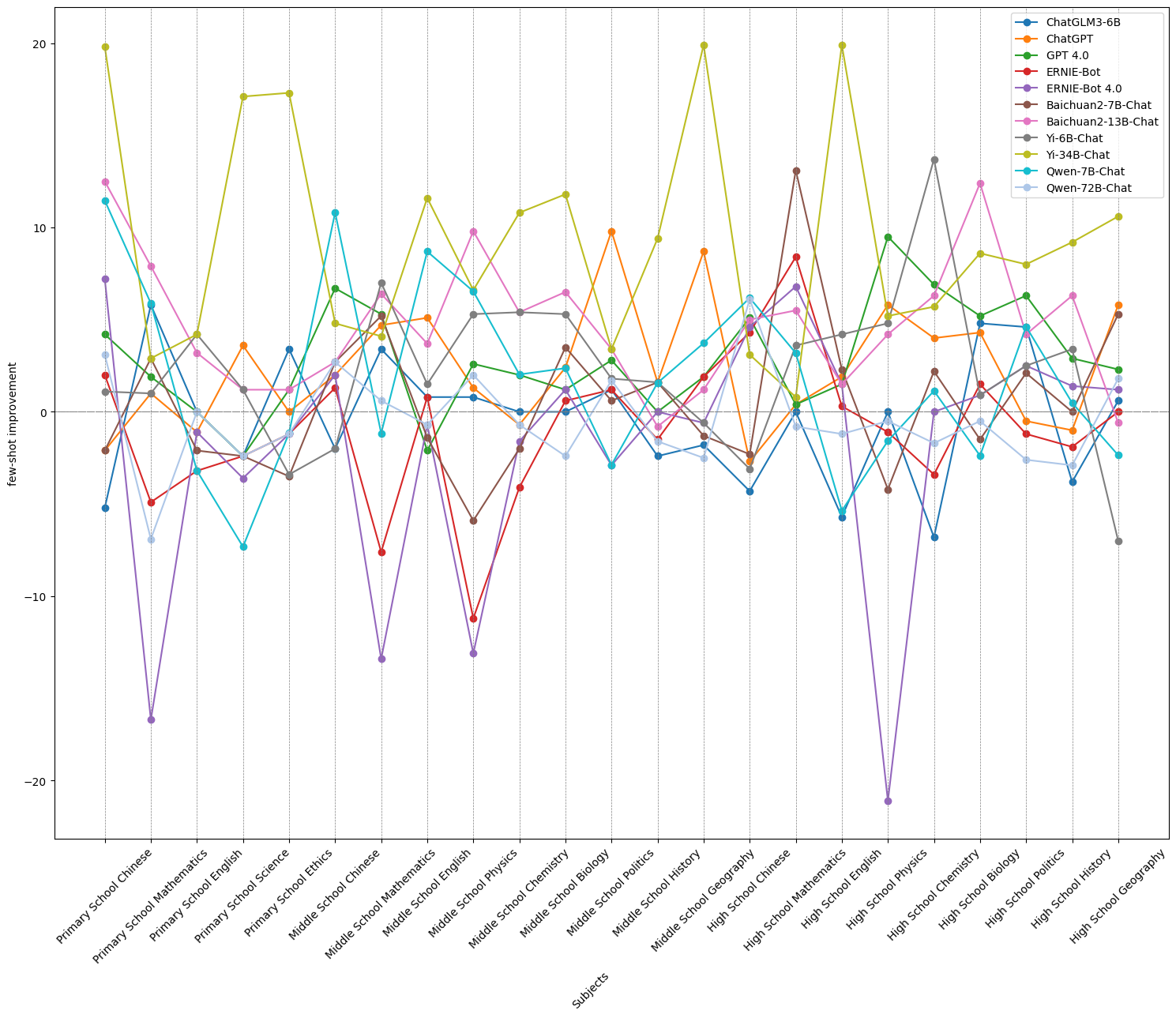}  
    \caption{Accuracy improvement from five-shot-ao to zero-shot across 23 subjects.}
    \label{fig:linefewshot} 
\end{figure*}

\textbf{Does Few-shot prompting help?} Table \ref{tab:model_performance_scenarios} and Figure \ref{fig:linefewshot} illustrate the performance of the models under zero-shot and five-shot conditions. In general, the accuracy of most models is slightly higher under five-shot conditions compared to zero-shot, which is expected. However, there is a decrease in performance for the top two models, Qwen-72B-Chat and ERNIE-Bot 4.0. This phenomenon of performance degradation with a limited number of examples has also been noted in other studies \citep{liu2023m3ke,zeng2023measuring,li2023cmmlu}. We believe that they can perform zero-shot reasoning without relying on Few-shot exemplars. Upon further observation from Table \ref{tab:3_arts_science_scenarios} and Figure \ref{fig:linefewshot}, it is evident that the improvement in arts subjects, especially in Chinese language, is more significant than that in science subjects. We speculate that there are two main reasons for this phenomenon: the quality of the sample and the richness of prior knowledge. The Arts test questions, which primarily required language comprehension and knowledge retention, were highly similar among the samples and were mostly helpful questions. On the contrary, science test questions are diverse and often involve formulas and calculations, so it is highly unlikely that the five samples will contain a sufficient number of useful samples. In addition, during the pre-training phase, the model is exposed to a large amount of textual data, and the arts test questions are formally more similar to the tasks that the model handles during this phase. As a result, the model has more prior knowledge related to the arts, and a small number of art samples can activate this relevant knowledge in time.

\begin{table}[ht]
  \centering
  \begin{tabularx}{\textwidth}{lXXXc}
    \toprule
    \textbf{Model} & \textbf{Zero-shot} & \textbf{Five-shot-ao} & \textbf{Five-shot-cot} & \textbf{Average} \\
    \midrule
    Random & 25.0 & 25.0 & 25.0 & 25.0 \\
    Qwen-72B-Chat & 89.0 & 88.7 & 88.8 & 88.9 \\
    Ernie-Bot 4.0 & 86.7 & 85.2 & 84.6 & 85.5 \\
    Yi-34B-Chat & 72.5 & 81.4 & 76.6 & 76.9 \\
    Ernie-Bot  & 76.1 & 75.7 & 75.7 & 75.9 \\
    GPT 4.0 & 70.5 & 73.8 & 67.4 & 70.6 \\
    Yi-6B-Chat & 68.8 & 71.2 & 66.5 & 68.8 \\
    Qwen-7B-Chat & 58.7 & 60.4 & 60.4 & 59.9 \\
    Baichuan2-13B-Chat & 56.1 & 60.9 & 56.1 & 57.8 \\
    ChatGLM3-6B & 59.8 & 59.2 & 53.7 & 57.6 \\
    Baichuan2-7B-Chat & 55.2 & 56.2 & 52.9 & 54.8 \\
    ChatGPT & 54.5 & 56.9 & 52.3 & 54.6 \\
    Chinese-Alpaca-2-13B & 44.8 & 46.2 & 38.7 & 43.3\\
    Educhat-sft-002-13B & 33.2 & 39.4 & 36.1 & 36.3 \\
    Chinese-LLaMA-2-13B & 35.7 & 38.9 & 33.2 & 35.9 \\
    Educhat-sft-002-13B-Baichuan & 54.0 & 14.4 & 38.1 & 35.5 \\
    \bottomrule
  \end{tabularx}
  \vspace{1em}
  \caption{Accuracy of multiple models in different prompt scenarios.}
  \label{tab:model_performance_scenarios}
\end{table}

\begin{figure*}[h]  
    \centering  
    \includegraphics[width=\textwidth]{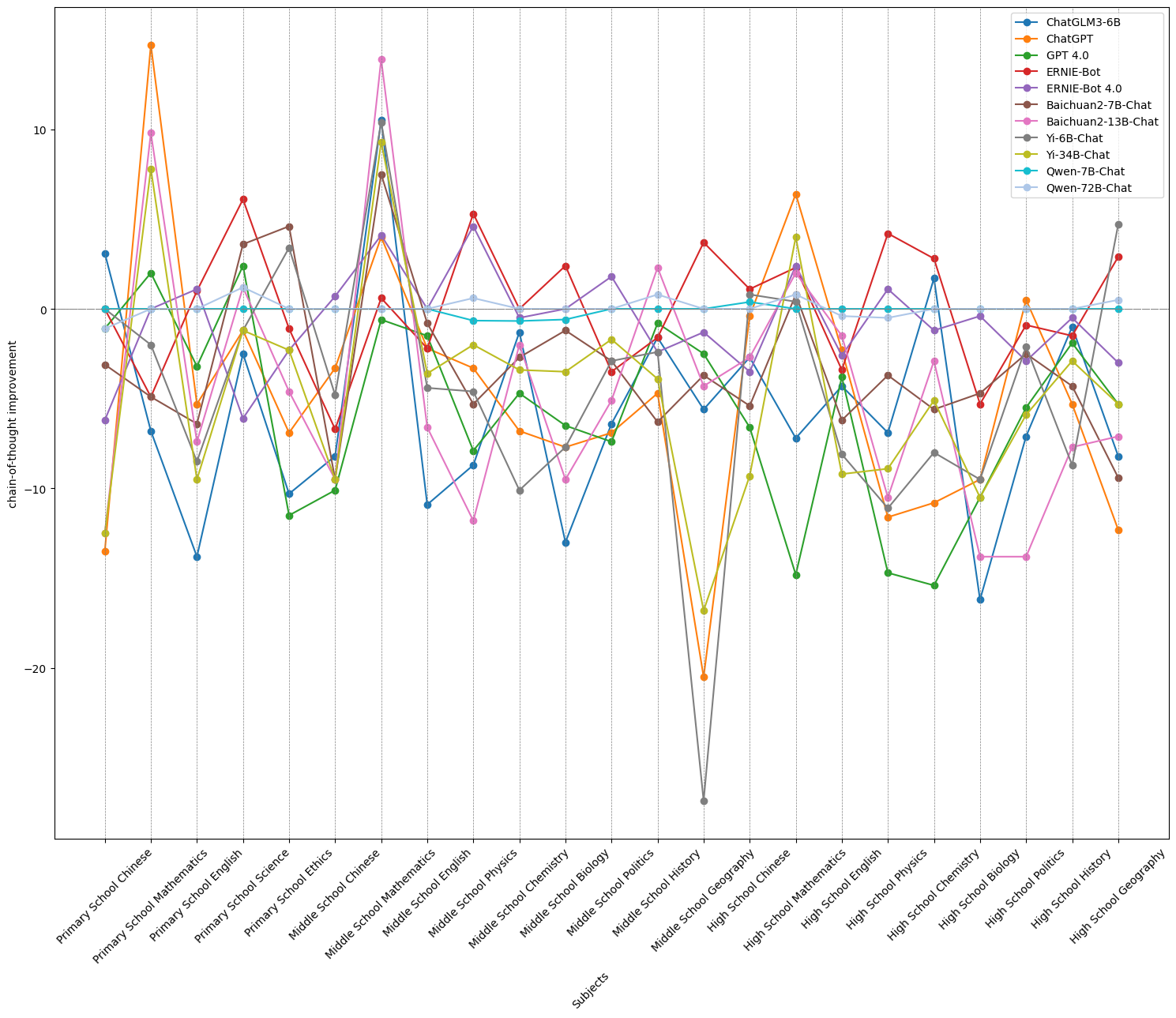}  
    \caption{Accuracy improvement from five-shot-ao to five-shot-cot across 23 subjects.}
    \label{fig:linecot}  
\end{figure*}

\begin{table}[ht]
  \centering
  \begin{tabularx}{\textwidth}{lccc}
    \toprule
    \textbf{Model} & \textbf{Arts} & \textbf{Science}  & \textbf{Average} \\
    \midrule
Qwen-72B-Chat                & 92.4 / 92.6 / 92.6 & 89.0 / 88.7 / 88.8 & 89.0 / 88.7 / 88.8 \\
ERNIE-Bot 4.0                & 90.5 / 91.8 / 90.1 & 86.7 / 85.2 / 84.6 & 86.7 / 85.2 / 84.6 \\
Yi-34B-Chat                  & 77.8 / 88.3 / 81.2 & 72.6 / 81.4 / 76.6 & 72.6 / 81.4 / 76.6 \\
ERNIE-Bot                    & 81.8 / 82.2 / 81.2 & 76.2 / 75.7 / 75.7 & 76.2 / 75.7 / 75.7 \\
GPT 4.0                      & 75.1 / 78.0 / 73.2 & 70.6 / 73.8 / 67.4 & 70.6 / 73.8 / 67.4 \\
Yi-6B-Chat                   & 76.1 / 76.5 / 71.7 & 68.8 / 71.3 / 66.6 & 68.8 / 71.3 / 66.6 \\
Qwen-7B-Chat                 & 65.9 / 68.1 / 68.1 & 58.8 / 60.5 / 60.4 & 58.8 / 60.5 / 60.4 \\
Baichuan2-13B-Chat           & 65.2 / 68.5 / 62.4 & 56.1 / 61.0 / 56.2 & 56.1 / 61.0 / 56.2 \\
ChatGLM3-6B                  & 64.8 / 63.4 / 57.9 & 59.8 / 59.3 / 53.8 & 59.8 / 59.3 / 53.8 \\
Baichuan2-7B-Chat            & 63.5 / 64.0 / 59.3 & 55.2 / 56.3 / 52.9 & 55.2 / 56.3 / 52.9 \\
ChatGPT                      & 61.0 / 63.1 / 57.5 & 54.6 / 56.9 / 52.4 & 54.6 / 56.9 / 52.4 \\
Chinese-Alpaca-2-13B         & 51.1 / 53.7 / 43.9 & 44.8 / 46.3 / 38.8 & 44.8 / 46.3 / 38.8 \\
5Educhat-sft-002-13B         & 39.5 / 46.3 / 39.9 & 33.3 / 39.4 / 36.1 & 33.3 / 39.4 / 36.1 \\
Chinese-LLaMA-2-13B          & 40.5 / 44.2 / 35.2 & 35.7 / 38.9 / 33.2 & 35.7 / 38.9 / 33.2 \\
Educhat-sft-002-13B-Baichuan & 60.0 / 15.9 / 44.9 & 54.1 / 14.4 / 38.1 & 54.1 / 14.4 / 38.1 \\
    \bottomrule
  \end{tabularx}
  \vspace{1em}
  \caption{Accuracy of multiple models in arts and science subjects across different prompt scenarios, the number on the left is in zero-shot, the number in the middle is in five-shot-ao, and the number on the right is five-shot-cot.}
  \label{tab:3_arts_science_scenarios}
\end{table}
\textbf{Does Chain-of-Thought prompting help?}
As observed in Table \ref{tab:model_performance_scenarios} and Figure \ref{fig:linecot}, compared to the five-shot-ao, nearly all models show a noticeable decline in average performance on the five-shot-cot. This observation aligns with the findings of \cite{huang2023c}, who noted a deterioration in model performance when applying CoT. We believe that many subjects in E-EVAL, especially in arts, do not require complex reasoning, and additional reasoning steps might decrease performance. Further observation from Table \ref{tab:3_arts_science_scenarios} and Figure \ref{fig:linecot} reveals that while the average performance decreased, there was a divergence between humanities and sciences, humanities showed a decline, whereas sciences, particularly high school and middle school mathematics, showed an increase. This is because science questions typically have fixed principles and a logical deduction process, hence CoT prompts can effectively guide models in structured reasoning. CoT can aid models in step-by-step construction of answers, which is advantageous for science questions. On the contrary, arts questions often involve broader and more ambiguous knowledge areas. These questions might depend more on intuition, experience, and understanding of polysemous terms, which are not suited for simple logical reasoning. Introducing CoT in arts questions could lead to models over-reasoning or developing reasoning chains in the wrong direction, as these questions might require a wider range of background knowledge and creative thinking, rather than simple step-by-step logical deduction. Therefore, the application of CoT needs to be adjusted based on the question type and complexity, to better adapt to the characteristics of different tasks.

\section{Related Work}
Initial benchmarking efforts were concentrated on simple, singular tasks such as part-of-speech tagging, and named entity recognition. However, with the continuous advancement of natural language processing technology, benchmarks have expanded to encompass more complex tasks, such as machine translation\citep{bojar-etal-2014-findings} and summarization\cite{narayan-etal-2018-dont,hermann2015teaching}. The advent of comprehensive benchmarks like GLUE\citep{wang2018glue} and SuperGLUE\citep{sarlin2020superglue} heralded a new era. These benchmarks amalgamate various natural language understanding tasks, including textual entailment, sentiment analysis, and question answering, thereby providing a standard for evaluating the holistic performance of models. Popular for their ability to assess models on both understanding and generating natural language, these benchmarks have gained prominence. Yet, the emergence of LLMs like BERT and GPT shifted the focus towards assessing performance on higher-level, more intricate tasks. Such models have even surpassed human-level performance on certain tasks, notably in text summarization\citep{hermann2015teaching} and reading comprehension\cite{rajpurkar2018know,li-etal-2022-multispanqa}. However, some recent work\citep{goyal2022news,liu2023gpteval} have demonstrate that LLM can perform even better than human or human annotators on some tasks such as summarization, leading to a re-evaluation of the appropriateness of using these benchmarks.To offer a more encompassing evaluation, new benchmarks like MMLU\citep{hendrycks2021measuring} include a multitude of domains and tasks, ranging from real-world exams to book knowledge, assessing ability in language understanding, common sense reasoning\citep{clark2018think,talmor-etal-2019-commonsenseqa,sakaguchi2021winogrande}, mathematical reasoning\citep{hendrycks2021measuring,cobbe2021training}, and code generation\citep{chen2021evaluating,austin2021program}. The BIG-bench\citep{srivastava2022beyond} includes 204 diverse tasks, some of which are deemed beyond the current ability of LLMs. The HELM\citep{liang2022holistic} benchmark comprises 42 distinct tasks, evaluating LLMs across seven metrics.\\
With the burgeoning development of Chinese Large Models, an increasing number of Chinese benchmarks have emerged. CLUE\citep{xu-etal-2020-clue}, an influential Chinese NLU benchmark, has been widely adopted in the field. Additionally, the team has recently introduced SuperCLUE\citep{xu2023superclue}, a benchmark tailored specifically for LLMs. Concurrently, Chinese benchmarks akin to MMLU\citep{hendrycks2021measuring} have surfaced, such as MMCUL\citep{zeng2023measuring}, which emphasizes medicine and education within its four major domains. AGIEval\citep{zhong2023agieval} focuses on standardized Chinese exams like the college entrance exam, while C-Eval\citep{huang2023c} encompasses questions across four levels of difficulty from middle school to professional tests. M3KE\citep{liu2023m3ke} gathers 71 tasks from the Chinese education examination system, akin to the coverage of C-Eval. CMMLU\citep{li2023cmmlu}, designed for the Chinese language and cultural context, is a fully localized Chinese benchmark.Compared to these benchmarks, E-EVAL distinguishes itself by (1) focusing on the field of K-12 Education in China, covering all subjects from primary to high school. (2) including the often-overlooked domain of elementary education. (3) sourcing data from homework and smaller-scale exams, ensuring a high degree of data privacy.
\section{Discussion and Conclusion}
Although large language models have potential in K-12 education, their accurate assessment remains critical for practical application, and the introduction of the E-EVAL benchmark, customized for K-12 education in China, provides a more accurate and comprehensive evaluation benchmark. Although E-EVAL is not a competitive ranking, it serves as a key tool for tracking the progress of large language models in Chinese K-12 education. This may pave the way for a wider and more effective utilization of large-scale language models in the field of K-12 education in China.

In this work, we obtained the following potentially insightful observations.
\begin{itemize}
    \item[$\bullet$] Chinese-dominant large language models have outperformed powerful Generalized models like GPT-4 in Chinese K-12 education.
    \item[$\bullet$] general, the more model parameters the better the results, but smaller models can also perform better than models of larger sizes.
    \item[$\bullet$] The gap between open-source and closed-source models is currently very tight, and open-source models are growing rapidly.
    \item[$\bullet$] Some specially trained educational large language models still lag behind generalized models in performance, suggesting that there is still much space for improvement in the education vertical.
    \item[$\bullet$] Overall, models perform slightly better in Few-shot compared to Zero-shot, with a more significant improvement observed in liberal arts subjects as opposed to science subjects.
    \item[$\bullet$]The application of CoT has a negative impact on the model as a whole, but it is helpful for complex science subjects such as mathematics.
    \item[$\bullet$] Large language models are better at liberal arts than science, and especially perform poorly in highly logical mathematics.
    \item[$\bullet$] The poor performance of the Chinese-dominant large language model on simple problems at primary school level may indicate that the model's mastery of higher-order knowledge does not mean that it has also mastered lower-order knowledge.
\end{itemize}

\section*{Acknowledgement}
This research was funded by the State Sponsored Postdoctoral Researcher Program of China. Grant No. GZC20232873. SIAT-UI AI and Education Joint Lab also funded this work.

\medskip
\bibliography{ref}

\appendix 

\section{Detailed Stats of E-EVAL}
\label{app:Detailed Stats of C-EVAL}
\subsection{Statistis of Subjects}
\label{app:statisticOfSbuject}
Table \ref{tab:summaryOfSubjects} lists all subjects of E-EVAL, their categories, and the number of questions for each subject.
\begin{table}[ht]
  \centering
  \begin{tabularx}{\textwidth}{lXc}
    \toprule
    \textbf{Subject} & \textbf{Category} & \textbf{\#Questions} \\
    \midrule
    Primary School Chinese (小学语文)     & Arts     & 96            \\
    Primary School Mathematics (小学数学) & Science  & 102           \\
    Primary School English (小学英语)     & Arts     & 94            \\
    Primary School Science (小学科学)     & Science  & 82            \\
    Primary School Ethics (小学道德)      & Arts     & 87            \\
    Middle School Chinese (初中语文)      & Arts     & 148           \\
    Middle School Mathematics (初中数学)  & Science  & 172           \\
    Middle School English (初中英语)      & Arts     & 138           \\
    Middle School Physics (初中物理)      & Science  & 152           \\
    Middle School Chemistry (初中化学)    & Science  & 148           \\
    Middle School Biology (初中生物)      & Science  & 169           \\
    Middle School Politics (初中政治)     & Arts     & 174           \\
    Middle School History (初中历史)      & Arts     & 127           \\
    Middle School Geography (初中地理)    & Arts     & 161           \\
    High School Chinese (高中语文)        & Arts     & 259           \\
    High School Mathematics (高中数学)    & Science  & 251           \\
    High School English (高中英语)        & Arts     & 261           \\
    High School Physics (高中物理)        & Science  & 190           \\
    High School Chemistry (高中化学)      & Science  & 176           \\
    High School Biology (高中生物)        & Science  & 210           \\
    High School Politics (高中政治)       & Arts     & 238           \\
    High School History (高中历史)        & Arts     & 207           \\
    High School Geography (高中地理)      & Arts     & 170           \\
    \bottomrule
  \end{tabularx}
  \vspace{1em}
  \caption{Summary of all 23 subjects.}
  \label{tab:summaryOfSubjects}
\end{table}
\subsection{Results of three prompt Evaluations}
\label{app:threeprompts}
The accuracy of zero-shot is shown in Table \ref{tab:model_performance_0shot}, The accuracy of five-shot-ao is shown in Table \ref{tab:model_performance_5shot}, The accuracy of five-shot-cot is shown in Table \ref{tab:model_performance_5shotcot}.

\subsection{Partial results per subject}
\label{app:resofsubjects}
Table \ref{tab:subject_performance} shows the performace of partial models on each subjects.
\begin{table}[ht]
  \centering
    \begin{adjustbox}{scale=0.85}
    
  \begin{tabular}{lccccc}
    \toprule
    \textbf{Subject} & \textbf{Qwen-72B-Chat} & \textbf{Ernie-Bot 4.0} & \textbf{Yi-34B} & \textbf{GPT 4.0} & \textbf{ChatGLM3-6B} \\
    \midrule
    \multicolumn{6}{c}{\textbf{Primary School}} \\
    Chinese & 87.5 / 90.6 / 89.5 & 87.5 / 94.7 / 88.5 & 66.6 / 86.4 / 73.9 & 70.8 / 75.0 / 73.9 & 56.2 / 51.0 / 54.1 \\
    Mathematics & 82.3 / 75.4 / 75.4 & 79.4 / 62.7 / 62.7 & 59.8 / 62.7 / 70.5 & 69.6 / 71.5 / 73.5 & 39.2 / 45.0 / 38.2 \\
    English & 95.7 / 95.7 / 95.7 & 96.8 / 95.7 / 96.8 & 90.4 / 94.6 / 85.1 & 92.5 / 92.5 / 89.3 & 62.7 / 62.7 / 48.9 \\
    Science & 90.2 / 87.8 / 89.0 & 92.6 / 89.0 / 82.9 & 68.2 / 85.3 / 84.1 & 85.3 / 82.9 / 85.3 & 71.9 / 69.5 / 67.0 \\
    Ethics & 97.7 / 96.5 / 96.5 & 97.7 / 96.5 / 94.2 & 80.4 / 97.7 / 95.4 & 94.2 / 95.4 / 83.9 & 83.9 / 87.3 / 77.0 \\
    \midrule
    \multicolumn{6}{c}{\textbf{Middle School}} \\
    Chinese & 85.8 / 88.5 / 88.5 & 87.1 / 89.1 / 89.8 & 75.6 / 80.4 / 70.9 & 54.7 / 61.4 / 51.3 & 52.0 / 50.0 / 41.8 \\
    Mathematics & 83.7 / 84.3 / 84.3 & 77.9 / 64.5 / 68.6 & 57.5 / 61.6 / 70.9 & 59.8 / 65.1 / 64.5 & 37.2 / 40.6 / 51.1 \\
    English & 97.8 / 97.1 / 97.1 & 97.8 / 97.1 / 97.1 & 83.3 / 94.9 / 91.3 & 93.4 / 91.3 / 89.8 & 70.2 / 71.0 / 60.1 \\
    Physics & 91.4 / 93.4 / 94.0 & 89.4 / 76.3 / 80.9 & 78.9 / 85.5 / 83.5 & 76.3 / 78.9 / 71.0 & 69.7 / 70.5 / 61.8 \\
    Chemistry & 95.2 / 94.5 / 94.5 & 91.2 / 89.6 / 89.1 & 75.0 / 85.8 / 82.4 & 67.5 / 69.5 / 64.8 & 67.5 / 67.5 / 66.2 \\
    Biology & 95.8 / 93.4 / 93.4 & 95.8 / 97.0 / 97.0 & 81.6 / 93.4 / 89.9 & 87.5 / 88.7 / 82.2 & 72.1 / 72.1 / 59.1 \\
    Politics & 94.2 / 95.9 / 95.9 & 94.2 / 91.3 / 93.1 & 92.5 / 95.9 / 94.2 & 83.9 / 86.7 / 79.3 & 82.7 / 83.9 / 77.5 \\
    History & 99.2 / 97.6 / 98.4 & 97.6 / 97.6 / 95.2 & 87.4 / 96.8 / 92.9 & 88.9 / 88.9 / 88.1 & 81.1 / 78.7 / 77.1 \\
    Geography & 97.5 / 95.0 / 95.0 & 95.6 / 95.0 / 93.7 & 75.7 / 95.6 / 78.8 & 81.9 / 83.8 / 81.3 & 70.1 / 68.3 / 62.7 \\
    \midrule
    \multicolumn{6}{c}{\textbf{High School}} \\
    Chinese & 83.0 / 89.1 / 89.1 & 74.9 / 79.5 / 76.0 & 62.9 / 66.0 / 56.7 & 39.3 / 44.4 / 37.8 & 40.5 / 36.2 / 33.5 \\
    Mathematics & 58.5 / 57.7 / 58.5 & 54.9 / 61.7 / 64.1 & 32.2 / 33.0 / 37.0 & 42.6 / 43.0 / 28.2 & 33.8 / 33.8 / 26.6 \\
    English & 95.0 / 93.8 / 93.4 & 94.2 / 95.7 / 93.1 & 72.0 / 91.9 / 82.7 & 88.5 / 90.0 / 86.2 & 64.7 / 59.0 / 54.7 \\
    Physics & 81.5 / 81.0 / 80.5 & 84.7 / 63.6 / 64.7 & 74.2 / 79.4 / 70.5 & 61.5 / 71.0 / 56.3 & 52.1 / 52.1 / 45.2 \\
    Chemistry & 91.4 / 89.7 / 89.7 & 86.9 / 86.9 / 85.7 & 68.7 / 74.4 / 69.3 & 59.0 / 65.9 / 50.5 & 51.1 / 44.3 / 46.0 \\
    Biology & 91.9 / 91.4 / 91.4 & 83.3 / 84.2 / 83.8 & 75.2 / 83.8 / 73.3  & 63.8 / 69.0 / 58.5 & 56.1 / 60.9 / 44.7 \\
    Politics & 94.1 / 91.5 / 91.5 & 88.6 / 91.1 / 88.2 & 82.7 / 90.7 / 84.8 & 65.5 / 71.8 / 66.3 & 70.1 / 74.7 / 67.6 \\
    History & 92.2 / 89.3 / 89.3 & 90.8 / 92.2 / 91.7 & 81.6 / 90.8 / 87.9 & 78.2 / 81.1 / 79.2 & 68.5 / 64.7 / 63.7 \\
    Geography & 88.2 / 90.0 / 90.5 & 88.8 / 90.0 / 87.0 & 73.5 / 84.1 / 78.8 & 78.8 / 81.1 / 75.8 & 57.0 / 57.6 / 49.4 \\
    \bottomrule
    
  \end{tabular}
  \end{adjustbox}
  \vspace{1em}
  \caption{Accuracy of partial models in all subjects across different prompt scenarios, the number on the left is in zero-shot, the number in the middle is in five-shot-ao, and the number on the right is in five-shot-cot.}
  \label{tab:subject_performance}
\end{table}

\begin{table}[ht]
  \centering
    \begin{tabularx}{\textwidth}{lXXXXXc}
    
      \toprule
      \textbf{Model} & \textbf{Arts} & \textbf{Science} & \textbf{Primary} & \textbf{Middle} & \textbf{High} & \textbf{Average} \\
      \midrule
Random                       & 25.0 & 25.0 & 25.0 & 25.0 & 25.0 & 25.0 \\
Qwen-72B-Chat                & 92.4 & 84.7 & 90.4 & 93.2 & 85.8 & 89.0 \\
ERNIE-Bot 4.0                & 90.5 & 81.7 & 90.4 & 91.6 & 82.4 & 86.7 \\
ERNIE-Bot                    & 81.8 & 68.9 & 80.0 & 82.3 & 70.9 & 76.2 \\
Yi-34B-Chat                  & 77.8 & 65.7 & 72.8 & 78.4 & 68.4 & 72.6 \\
GPT 4.0                      & 75.1 & 64.7 & 82.0 & 76.8 & 63.5 & 70.6 \\
Yi-6B-Chat                   & 76.1 & 59.3 & 71.3 & 76.2 & 63.0 & 68.8 \\
ChatGLM3-6B                  & 64.8 & 53.4 & 61.8 & 66.6 & 54.6 & 59.8 \\
Qwen-7B-Chat                 & 65.9 & 49.5 & 68.8 & 63.8 & 52.9 & 58.8 \\
Baichuan2-13B-Chat           & 65.2 & 44.3 & 66.5 & 63.4 & 48.6 & 56.1 \\
Baichuan2-7B-Chat            & 63.5 & 44.4 & 62.6 & 61.9 & 48.8 & 55.2 \\
ChatGPT                      & 61.0 & 46.1 & 68.9 & 57.5 & 49.1 & 54.6 \\
Educhat-sft-002-13B-Baichuan & 60.0 & 46.2 & 60.3 & 56.5 & 50.9 & 54.1 \\
Chinese-Alpaca-2-13B         & 51.1 & 36.7 & 54.2 & 48.3 & 40.2 & 44.8 \\
Chinese-LLaMA-2-13B          & 40.5 & 29.4 & 34.7 & 38.3 & 34.1 & 35.7 \\
5Educhat-sft-002-13B         & 39.5 & 25.1 & 37.3 & 36.9 & 29.7 & 33.3 \\
      \bottomrule
    \end{tabularx}
    \vspace{1em}
  \caption{Zero-shot accuracy of multiple models in answer-only setting.}
  \label{tab:model_performance_0shot}
\end{table}

\begin{table}[ht]
  \centering
    \begin{tabularx}{\textwidth}{lXXXXXc}
    
      \toprule
      \textbf{Model} & \textbf{Arts} & \textbf{Science} & \textbf{Primary} & \textbf{Middle} & \textbf{High} & \textbf{Average} \\
      \midrule
Random                       & 25.0 & 25.0 & 25.0 & 25.0 & 25.0 & 25.0 \\
Qwen-72B-Chat                & 92.6 & 83.7 & 88.9 & 93.1 & 85.6 & 88.7 \\
ERNIE-Bot 4.0                & 91.8 & 76.6 & 87.1 & 88.2 & 82.7 & 85.2 \\
Yi-34B-Chat                  & 88.3 & 72.4 & 84.8 & 87.4 & 76.4 & 81.4 \\
ERNIE-Bot                    & 82.2 & 67.3 & 78.0 & 80.2 & 72.0 & 75.7 \\
GPT 4.0                      & 78.0 & 68.4 & 83.0 & 79.2 & 67.8 & 73.8 \\
Yi-6B-Chat                   & 76.5 & 64.4 & 72.2 & 79.1 & 65.5 & 71.3 \\
Baichuan2-13B-Chat           & 68.5 & 51.2 & 72.0 & 67.8 & 53.6 & 61.0 \\
Qwen-7B-Chat                 & 68.1 & 50.5 & 70.3 & 67.1 & 53.5 & 60.5 \\
ChatGLM3-6B                  & 63.4 & 54.0 & 62.2 & 66.7 & 53.4 & 59.3 \\
ChatGPT                      & 63.1 & 48.9 & 69.1 & 61.5 & 50.8 & 56.9 \\
Baichuan2-7B-Chat            & 64.0 & 46.2 & 61.3 & 62.3 & 50.8 & 56.3 \\
Chinese-Alpaca-2-13B         & 53.7 & 36.6 & 54.0 & 50.4 & 41.5 & 46.3 \\
5Educhat-sft-002-13B         & 46.3 & 30.4 & 44.0 & 41.0 & 37.2 & 39.4 \\
Chinese-LLaMA-2-13B          & 44.2 & 31.9 & 45.9 & 43.1 & 34.3 & 38.9 \\
Educhat-sft-002-13B-Baichuan & 15.9 & 12.5 & 17.8 & 18.8 & 10.6 & 14.4 \\
      \bottomrule
    \end{tabularx}
    \vspace{1em}
  \caption{Five-shot accuracy of multiple models in answer-only setting.}
  \label{tab:model_performance_5shot}
\end{table}

\begin{table}[ht]
  \centering
    \begin{tabularx}{\textwidth}{lXXXXXc}
    
      \toprule
      \textbf{Model} & \textbf{Arts} & \textbf{Science} & \textbf{Primary} & \textbf{Middle} & \textbf{High} & \textbf{Average} \\
      \midrule
Random                       & 25.0 & 25.0 & 25.0 & 25.0 & 25.0 & 25.0 \\
Qwen-72B-Chat                & 92.6 & 83.8 & 88.9 & 93.3 & 85.7 & 88.8 \\
ERNIE-Bot 4.0                & 90.1 & 77.4 & 84.6 & 89.1 & 81.5 & 84.6 \\
Yi-34B-Chat                  & 81.2 & 70.6 & 81.3 & 83.7 & 70.5 & 76.6 \\
ERNIE-Bot                    & 81.2 & 68.6 & 78.0 & 80.1 & 72.1 & 75.7 \\
GPT 4.0                      & 73.2 & 59.8 & 80.9 & 74.5 & 59.3 & 67.4 \\
Yi-6B-Chat                   & 71.7 & 59.9 & 70.5 & 73.2 & 61.0 & 66.6 \\
Qwen-7B-Chat                 & 68.1 & 50.3 & 70.3 & 66.8 & 53.6 & 60.4 \\
Baichuan2-13B-Chat           & 62.4 & 48.1 & 69.4 & 64.2 & 47.4 & 56.2 \\
ChatGLM3-6B                  & 57.9 & 48.5 & 56.1 & 61.8 & 47.6 & 53.8 \\
Baichuan2-7B-Chat            & 59.3 & 44.5 & 59.8 & 59.8 & 46.4 & 52.9 \\
ChatGPT                      & 57.5 & 45.7 & 67.0 & 55.7 & 46.6 & 52.4 \\
Chinese-Alpaca-2-13B         & 43.9 & 32.0 & 46.2 & 41.5 & 35.1 & 38.8 \\
Educhat-sft-002-13B-Baichuan & 44.9 & 29.3 & 44.5 & 40.2 & 35.2 & 38.1 \\
5Educhat-sft-002-13B         & 39.9 & 31.2 & 38.4 & 42.0 & 31.4 & 36.1 \\
Chinese-LLaMA-2-13B          & 35.2 & 30.7 & 37.3 & 34.4 & 31.5 & 33.2 \\
      \bottomrule
    \end{tabularx}
    \vspace{1em}
  \caption{Five-shot accuracy of multiple models in Chain-of-Thought setting.}
  \label{tab:model_performance_5shotcot}
\end{table}

\section{Models being Evaluated}
\label{app:models}
\textbf{Baichuan 2-13B and Baichuan 2-7B} are the new generation of open-source large language models launched by Baichuan Intelligence. It is trained on a high-quality corpus with 2.6 trillion tokens and has achieved the best performance in authoritative Chinese and English benchmarks of the same size.Baichuan 2 comes in two model variants: Baichuan 2-7B with 70 billion parameters and Baichuan 2-13B with 130 billion parameters. Both models have undergone training on a massive 26 trillion tokens. In this paper, we evaluate the models Baichuan 2-7B-Chat and Baichuan 2-13B-Chat, specifically optimized for adhering to human instructions. These models demonstrate outstanding performance in dialogue and context comprehension.\\[1ex]
\textbf{Qwen-72B and Qwen-7B} are integral components of the Qwen series of language models developed by Alibaba Cloud. Both models are built upon the Transformer architecture and have been trained on a diverse range of data sources, including internet texts, professional literature, and code. Qwen-72B, boasting a substantial 72 billion parameters, excels in multiple Chinese and English downstream tasks, particularly in areas such as reasoning and translation. It has undergone extensive pretraining on over 3 trillion tokens, encompassing a wide array of languages and domains, and can support contexts of up to 32,000 tokens in length.On the other hand, the 7-billion-parameter Qwen-7B also demonstrates remarkable data coverage and diversity. In this paper, the models evaluated are the chatbot variants fine-tuned from Qwen-72B and Qwen-7B through human alignment techniques. \\[1ex]
\textbf{ChatGLM3-6B} is the latest open-source model in the ChatGLM series, excels with its base model, ChatGLM3-6B-Base, incorporating diverse training datasets, sufficient training steps, and effective strategies. This culminates in superior performance on various datasets, including those involving semantics, mathematics, reasoning, coding, and knowledge, especially notable among models with less than 10 billion parameters. The model introduces an innovative Prompt format, enhancing multi-turn dialogues, function calls, code interpretation, and agent tasks. It represents a significant advance in bilingual (Chinese and English) language processing, particularly in question answering and dialogue tasks.\\[1ex]
\textbf{Yi-34B and Yi-6B} are two large language models in the Yi series developed by 01.AI. Trained from scratch on a 3T multilingual corpus, they exhibit exceptional bilingual ability, excelling in language comprehension, commonsense reasoning, and reading comprehension. Yi-34B performed outstandingly in various assessments, ranking second only to GPT 4.0 on the AlpacaEval leaderboard in December 2023, surpassing major models like LLaMA2-Chat-70B. In the field of Chinese, it ranked second in the SuperCLUE in October 2023, again only behind GPT 4.0, and ahead of models like Baidu's ERNIE. Although Yi-6B has fewer parameters than Yi-34B, it plays a crucial role in innovative projects and diverse applications, demonstrating strong language processing ability. In this paper, we are using the chat versions of Yi-34B and Yi-6B.\\[1ex]
\textbf{ChatGPT and GPT 4.0}, developed by OpenAI, represent the latest advancements in the GPT series of AI models. While ChatGPT is based on ChatGPT and focuses on smooth conversational experiences, GPT 4.0 is the newest iteration, introducing the capability to process visual inputs, enriching user interactions with the model. GPT 4.0 has demonstrated improved factual response rates and a significant reduction in responses to inappropriate content in OpenAI's internal tests. These models are trained to better follow human instructions, offering helpfulness, harmlessness, and honesty. GPT 4.0's updates also include an increased input/output capacity, enhanced creativity and collaborative ability, and the ability to connect to third-party knowledge sources.\\[1ex]
\textbf{ERNIE-Bot and ERNIE-Bot 4.0} are advanced large language models developed by Baidu. ERNIE-Bot is an industrial-grade, knowledge-enhanced large language model that encompasses extensive Chinese data. It possesses robust capabilities in text comprehension, dialogue question-answering, and content creation. The 4.0 version of ERNIE-Bot represents a comprehensive upgrade of the foundational model, achieving significant improvements in understanding, generation, logic, and memory capabilities compared to its predecessor. The notable feature of ERNIE-Bot 4.0 is its multimodal capability, capable of generating a range of content including text, images, and videos based on simple text prompts and image inputs. Compared to ChatGPT, ERNIE-Bot 4.0 has an advantage in multimodal ability, though ChatGPT Plus with GPT 4.0 provides multimodality, it currently does not support video generation.\\[1ex]
\textbf{Chinese LLaMA  \& Alpaca} LLM project is based on the LLaMA-2, released by Meta. Developers open-source Chinese LLaMA-2 (foundation model) and Alpaca-2 (instruction-following model). 
These models extend the original LLaMA-2 structure by incorporating an additional 20,000 Chinese tokens into its vocabulary and undergoes secondary pre-training and instruction fine-tuning on Chinese data, which further improved the fundamental semantic understanding of the Chinese language, resulting in a significant performance improvement compared to the first-generation models.In this paper, we used Chinese-LLaMA-2-13B and Chinese-Alpaca-2-13B.\\[1ex]
\textbf{EduChat} is a large-scale language model (LLM)-based chatbot system in the education domain. Its goal is to support personalized, fair, and compassionate intelligent education, serving teachers, students, and parents. Guided by theories from psychology and education, it further strengthens educational functions such as open question answering, essay assessment, Socratic teaching, and emotional support based on the existing basic LLMs.Developers use an educational corpus for pre-training to enable the model to acquire domain-specific knowledge. They further fine-tune the model on designed system prompts and instructions to stimulate a range of tool usage skills.They proposed several versions of the model. In this paper, we evaluated two models in educhat, educhat-sft-002-13B-Baichuan and educhat-sft-002-13B.\\[1ex]



\end{CJK}
\end{document}